%% file: shifter-quantization.tex
\documentclass[11pt]{article}
\usepackage[margin=1in]{geometry}
\usepackage[T1]{fontenc}
\usepackage{lmodern,microtype}
\usepackage{amsmath,amssymb,booktabs,tabularx,array,flafter}
\usepackage{graphicx,float,color}
\usepackage[numbers,sort&compress]{natbib}
\usepackage[hidelinks]{hyperref}
\newcolumntype{L}{>{\raggedright\arraybackslash}X}
\newcolumntype{R}{>{\raggedleft\arraybackslash}X}
\newcommand{\clip}{\operatorname{clip}}
\newcommand{\round}{\operatorname{round}}

\title{Shifter Quantization: Compact Representations for Native Integer Computation}
\author{Yufeng Huang \\ \texttt{yufeng.huang@columbia.edu}}
\date{}
\input{results/serving-tables.tex}
\input{results/model-quality.tex}
\input{results/sq8-evaluation.tex}
\input{results/shi2-evaluation.tex}
\input{projection/2bit/benchmark-rows.tex}
\input{projection/4bit/benchmark-rows.tex}
\begin{document}
\maketitle

\begin{abstract}
Groupwise quantization is used in LLM quantization to improve accuracy at low bit widths,
but grouping weights keeps standard integer MMA from spanning the full dot product for
efficient GPU inference. We develop a new quantization method, Shifter Quantization
(SQ), that reconstructs native-width integer operands by appending and shifting
shared bits. To further extend the efficiency, we develop a special variant of SQ,
Shifter Integer (SHI), to use INT4 kernels. We show that SQ and SHI formats
perform better than standard groupwise quantization in quality, as measured by projection error
and KL divergence, and in inference speed, as measured during decoding and prefill.
For low precision like 2-bit quantization, SQ2 and SHI2 achieve $2.71$--$3.77\times$ the
4,096-token prefill throughput of llama.cpp on a Q2\_K
checkpoint and reduce WikiText-2 KL by 6.6--7.8\%. For mid precision like 4-bit quantization,
SQ4 achieves $1.43$--$2.47\times$ the prefill throughput of NInfer's
groupwise-integer baselines across the tested models and prompt lengths,
with 24--40\% lower KL than the NInfer baselines. For high precision like 8-bit quantization,
SQ8 achieves \WEightAEightSpeedGain\% higher batch-one prefill
throughput and \WEightAEightKLGain\% lower KL than vLLM's online FP8.
\end{abstract}

\begin{center}
\small
\textbf{Results at a glance}\par
\smallskip
\begin{tabularx}{\linewidth}{@{}l L r r@{}}
\toprule
\textbf{Shifter format} & \textbf{Baseline} & \textbf{Prefill throughput} & \textbf{Lower WikiText-2 KL} \\
\midrule
2-bit SQ2 and SHI2 & llama.cpp Q2\_K & $2.71$--$3.77\times$ & 6.6--7.8\% \\
4-bit SQ4 & NInfer groupwise integer & $1.43$--$2.47\times$ & 24--40\% \\
8-bit SQ8 & vLLM online FP8 & \WEightAEightSpeedGain\% higher & \WEightAEightKLGain\% \\
\bottomrule
\end{tabularx}
\end{center}

\section{Introduction}
\label{sec:introduction}

As large language models continue to grow, frontier models now
exceed one trillion parameters~\citep{kimik3,qwen38max,deepseekv4,longcat20}.
Local inference typically uses smaller models, but consumer and
workstation GPUs still face tight memory budgets. Model weights share
that memory with the attention key-value (KV) cache, which grows with
context length and the number of active requests~\citep{vllm,kivi}.

Quantizing weights and the KV cache helps fit larger models and longer
contexts on local hardware~\citep{llmint8,kivi}. At low bit widths, however, quantization
must balance the range needed for large weights against the resolution
needed for small ones. Groupwise scaling addresses this trade-off by
assigning scales to groups of values. It is used in academic methods
such as GPTQ and AWQ~\citep{gptq,awq}, llama.cpp's Qx\_0, Qx\_K, and
IQx families~\citep{llamacpp}, and hardware-supported formats such as
NVFP4 and MXFP4 on NVIDIA Blackwell and MXFP4 on AMD
CDNA~4~\citep{nvfp4,mxspec,nvidiaptx,amdcdna4}.

Groupwise quantization also affects inference speed. When a dot product
spans groups with different scales, the kernel must preserve each
group's contribution separately so that it can apply the corresponding
scale~\citep{atom,qserve}. This prevents conventional integer matrix multiply--accumulate
(MMA) instructions from accumulating across the full reduction
dimension. A kernel can use MMA within each group, but the small
groups limit how effectively it can use the hardware~\citep{atom,qserve}. Recent GPUs
provide block-scaled instructions for a limited set of formats and
group sizes~\citep{nvidiaptx,amdcdna4}. Supporting other groupwise representations efficiently
remains a challenge.

To address this limitation, we introduce Shifter Quantization (SQ) to make
low-bit local inference more efficient by enabling MMA across the full
reduction dimension. Before multiplication, we expand weights stored at low
precision in GPU registers using cheap bit shifting operations. We apply these
operations to groups of four or eight weights (G4 or G8), smaller than the G16,
G32, or G64 groups commonly used in conventional quantization. Smaller groups
let the representation adapt to weight variation at a finer scale, which can
reduce quantization error. SQ preserves the finer adaptation of small groups
while allowing integer MMA to span the full reduction dimension.

Building on SQ, we develop Shifter Integer (SHI) to take advantage of
faster INT4 kernels. SHI keeps the main product in INT4 and applies group
corrections through shorter products when needed. With groups of eight,
each correction product has one eighth as many terms as the main product.

\begin{itemize}
\item \textbf{Shifter Quantization (SQ).} SQ stores low-bit weight codes
with shared information for each group. It reconstructs integer operands
by appending shared bits, applying a group-specific shift, and truncating
to the target width. The reconstructed operands can use integer MMA
across the full reduction dimension, with channel scaling applied afterward.
\item \textbf{Shifter Integer (SHI).} SHI builds on SQ to use INT4 for
the main matrix product. It represents group-level weight corrections
with additional integer products over group sums. These products are
shorter than the main product and can also account for activation
corrections.
\item \textbf{Implementation and evaluation.} We evaluate SQ and SHI for
local inference, measuring projection error, full-model quality, and serving
throughput. We compare two-, four-, and eight-bit checkpoints with
llama.cpp~\citep{llamacpp}, NInfer~\citep{ninfer}, and vLLM~\citep{vllm},
respectively, and measure SQ4 prefill and decode throughput across four GPUs.
\end{itemize}

Section~\ref{sec:methods} defines the representations and execution
profiles. Section~\ref{sec:projection} isolates one MLP projection to show
the effects of the execution paths on latency and error.
Section~\ref{sec:full-model-quality} evaluates full checkpoints against BF16
references and measures prefill. Section~\ref{sec:serving} compares the
four-bit serving systems on prefill and speculative decode.
Section~\ref{sec:cross-gpu} measures how SQ4 prefill and decode throughput
scale across four GPUs.

\section{Background and related work}
\label{sec:background}

Groupwise scaling gives each group a range suited to its values. The
idea predates neural networks and appears in block-floating-point
arithmetic for digital filters~\citep{oppenheim}. VS-Quant studied
finer-grained scaling for neural network inference~\citep{vsquant}.
Smaller groups can represent local ranges more accurately. They also require
more scale metadata and make the kernel apply scales more often.

Several low-bit methods use groupwise or blockwise scaling. GPTQ and
AWQ support groupwise integer weight quantization~\citep{gptq,awq};
ZeroQuant combines fine-grained quantization with optimized inference
kernels~\citep{zeroquant}.
QLoRA's NF4 combines block scales with nonuniform element
values~\citep{qlora}, while llama.cpp provides Q and IQ formats for
low-bit inference~\citep{llamacpp}. These formats divide their storage
among weight codes, scales, and other metadata. We report packed bits
per weight alongside accuracy and speed in the projection benchmarks.

Group scales also affect how a GPU computes a dot product. For weight
codes $q_k$, group scales $s_g$, and activation values $\xi_k$, an output
is $\sum_g s_g\sum_{k\in g}q_k \xi_k$. When scales differ across groups,
each partial sum must be scaled before the results are combined. Atom
uses integer MMA to compute group products and applies their scales
within the kernel~\citep{atom}. QServe identifies weight and partial-sum
dequantization as substantial sources of overhead in low-bit GPU
inference~\citep{qserve}. SQ instead incorporates group information
into reconstructed integer operands. SHI expresses group corrections
as additional integer products over group sums.

Block-scaled floating-point formats combine local scales with
hardware-supported arithmetic~\citep{sharedmx}. MXFP4 and MXFP8 share an
exponent scale across 32 elements~\citep{mxpaper,mxspec}. NVFP4 uses an
FP8 scale per 16 four-bit elements and a global scale~\citep{nvfp4}.
Recent GPU instructions support specific block-scaled formats and
group sizes~\citep{nvidiaptx,amdcdna4}. Because SHI4 and NVFP4 use similar amounts of weight storage, we compare
their accuracy and speed.

Outliers make low-bit quantization harder.
LLM.int8() isolates outlier features in a higher-precision computation~\citep{llmint8}.
SmoothQuant redistributes quantization difficulty across activation and
weight channels~\citep{smoothquant}. QuaRot uses rotations to spread
outliers across channels~\citep{quarot}. We use channel scales and
signed Hadamard rotation to condition the inputs and weights. SQ and
SHI then determine how the transformed weights are stored and used in
integer computation.

\section{Quantization formats and execution}
\label{sec:methods}

\subsection{Linear projections and the quantization basis}

Consider a linear projection with row-oriented activations $X\in\mathbb{R}^{M\times K}$ and weights $W\in\mathbb{R}^{N\times K}$:
\begin{equation}
    Y=XW^\top.
\end{equation}
Here $M$ denotes the number of input tokens, $K$ the reduction dimension, and $N$ the output width. We quantize linear projections while retaining the surrounding attention, recurrent-state and nonlinear computation. Different projections may use different representations; the bit width assigned to an MLP bank therefore does not describe the average precision of the complete model.

We condition each projection using an orthogonal transform $U$ and diagonal channel scales $C_i$ and $C_o$. An integer reconstruction $P$ is fitted such that
\begin{equation}
    WU\simeq C_o P C_i,
    \qquad
    Y\simeq (XU C_i)P^\top C_o.
    \label{eq:basis}
\end{equation}
The first relation defines the fitted weight basis, and the second follows from $UU^\top=I$. Input-channel scales act before the contraction; output-channel scales act after it. Projections fitted independently may have different $C_i$ even when they share an input, so fused execution retains their separate scale vectors.

Our block transform is a signed, normalized Hadamard rotation. For blocks of $B=512$ input channels, write
\begin{equation}
    U_B=D_B H_B, \qquad
    H_B H_B^\top=I,
\end{equation}
where $D_B$ contains the signs and $H_B$ is the normalized Sylvester Hadamard matrix in natural binary order. The full transform $U$ is block diagonal. Row-oriented inputs receive the signs, the Hadamard butterfly and then input-channel scaling; the inverse transform reverses the transform order. Explicitly stored signs make the rotation reproducible across backends. We apply it at linear boundaries without moving it through nonlinearities. Rotation is an established conditioning technique~\citep{quarot}; it conditions the quantization basis independently of the SQ and SHI code constructions.

Throughout the method, we distinguish stored codes from reconstructed operands and scaled weights. A three-bit code can expand to an INT8 operand while remaining compact in storage, although retaining the expanded bank increases resident memory. Memory accounting therefore separates code bits, channel parameters, transform data, prepared weight planes and temporary workspace.

\subsection{SQ integer reconstruction}
\label{sec:sq}

SQ expands compact codes into INT4 or INT8 operands through three bitwise operations: append shared bits, apply a grouped shift, and truncate to the target width. These integer operations incorporate group-level information before multiplication, replacing floating-point groupwise reconstruction. The complete path is
\begin{align*}
    z &= (q\mathbin{\ll}m)+a_g
       &&\text{(1) append shared bits},\\
    v &= (z\mathbin{\ll}\ell)\mathbin{\gg}r_g
       &&\text{(2) apply the grouped shift},\\
    P &= \clip_t(v\mathbin{\gg}\delta)
       &&\text{(3) truncate to the target width}.
\end{align*}
Here $q$ is the signed base code, $a_g$ and $r_g$ are shared group fields, and $t$ is the native operand width. The symbols $\ll$ and $\gg$ denote integer left shift and arithmetic right shift, with signed rounding defined below. Fixed alignment $\ell$ and final reduction $\delta$ determine the retained bit positions; they add no learned group fields. Appending expands the code, shifting supplies local power-of-two adaptation, and truncation produces the target-width operand. Local adaptation is thus incorporated during integer preparation, leaving channel scaling and output conversion outside the integer reduction.

We name an SQ format SQ$x$-G$y$A$m$S$n$: $x$ is the width of each base code $q$, $y$ is the number of weights per group, $m$ is the width of the shared append field $a_g$, and $n$ is the width of the group shift selector $r_g$. For example, SQ2-G4A1S1 uses two-bit codes in groups of four, with one append bit and a one-bit shift selector per group. We describe the three operations below.

\subsubsection{Append shared bits}
Let $q_{j,k}$ be a signed $x$-bit code for output channel $j$ and input channel $k$. Partition each weight row into groups of $y$ consecutive input channels, with $g(k)=\lfloor k/y\rfloor$. Each group stores an $m$-bit append field $a_{j,g}$. The expanded integer is
\begin{equation}
    z_{j,k}=2^m q_{j,k}+a_{j,g(k)},
    \quad
    \begin{cases}
    -2^{x-1}\le q_{j,k}<2^{x-1},\\
    0\le a_{j,g}<2^m.
    \end{cases}
    \label{eq:append}
\end{equation}
The left shift opens $m$ low-order positions, which are filled by the shared field. This expands each code without storing all appended bits per element. For example, an INT4 base with four appended bits spans the INT8 range, while sharing those bits across $y=8$ weights adds only $4/8=0.5$ bits per weight. Shift fields and channel parameters contribute separately to storage.

\subsubsection{Apply the grouped shift}
A shared selector $r_{j,g}$ positions the expanded code within the operand. Before integer rounding, a right shift by $r$ scales the value by $2^{-r}$: shifts of one, two and three positions correspond to factors $1/2$, $1/4$ and $1/8$. Grouped shifting therefore provides local scale adaptation using integer instructions rather than floating-point multiplication.

\subsubsection{Truncate to the hardware width}
The final step reduces the positioned value to a signed $t$-bit operand. Define
\begin{equation}
    R_d(v)=\left\lfloor v/2^d\right\rfloor,
    \qquad d\ge0,
\end{equation}
as arithmetic right shift, including its rounding for negative inputs. Combining the three steps gives
\begin{equation}
    P_{j,k}=\clip_t\!\left(
       R_\delta\!\left(R_{r_{j,g(k)}}(2^{\ell}z_{j,k})\right)
    \right),
    \label{eq:sq}
\end{equation}
where $\ell\ge0$ specifies alignment, $\delta\ge0$ specifies the final reduction, and $\clip_t$ applies saturation if required. These parameters and the saturation rule are part of the representation. Saturation is inactive when all reduced values fit. Before rounding, the effective factor is $2^{\ell-r-\delta}$; fixed normalization is absorbed into the channel scales.

The same effective shift can be indexed in the opposite direction: a nonnegative expansion selector $s$ corresponds to $r=\ell-s$ when the selector ranges permit. Either convention specifies the same reconstruction only if rounding and range handling also agree. Truncation discards low-order bits after sufficiently wide intermediate computation; retaining the low byte of an overflowing value would instead cause wraparound.

Figure~\ref{fig:sq4-reconstruction} shows the bit operations for SQ4-G8A2S2.
A two-bit example is given in Appendix~\ref{app:sq-example}.

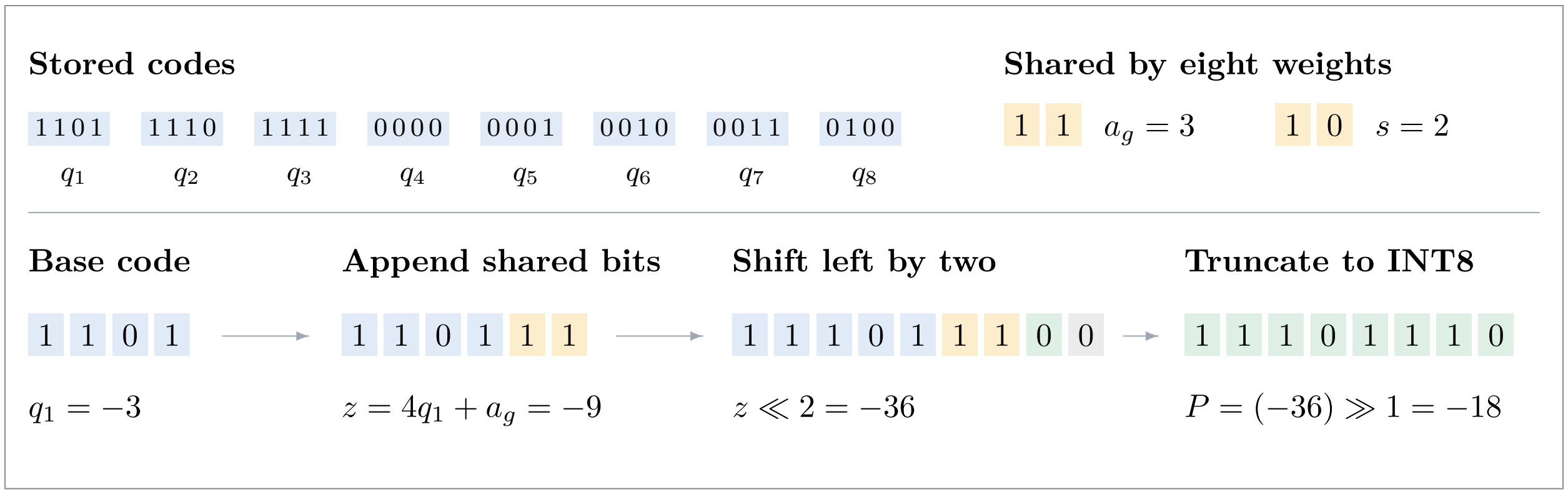
\begin{figure}[!htbp]
\centering
\setlength{\fboxsep}{7pt}
\setlength{\fboxrule}{0.4pt}
\resizebox{\linewidth}{!}{\fcolorbox[gray]{0.55}{1}{\input{figures/sq4-g8a2s2.tex}}}
\caption{SQ4-G8A2S2 reconstructs a signed INT8 operand from a four-bit
base code and two two-bit fields shared by eight weights, for
$4+(2+2)/8=4.5$ packed bits per weight (bpw). The stored left-shift
selector $s=2$ corresponds to right-shift count $r=1$ from full alignment
($\ell=3$). The final reduction drops one low bit. At output scale $0.1$,
append bits $11$ give $-1.80$ for the first weight, and append bits $00$
give $-2.40$ with the same shift.}
\label{fig:sq4-reconstruction}
\end{figure}

\subsubsection{Storage and arithmetic}
For complete groups, the code storage is
\begin{equation}
    B_{\mathrm{SQ}}=x+\frac{m+n}{y}
    \quad\text{bits per weight}.
    \label{eq:sqbpw}
\end{equation}
Channel scales, optional offsets, transform signs and padding are additional. Signedness, shift interpretation and range handling complete the descriptor by specifying Equation~\eqref{eq:sq}.

SQ2--SQ6 primarily reconstruct INT8 operands. A tiled implementation unpacks and shifts codes into compute fragments or bounded staging buffers, then executes INT8 MMA with INT32 accumulation. Because $P$ already incorporates local adaptation, the dot product requires no floating-point group scale along $K$. Power-of-two scaling and shared exponents are established techniques~\citep{sharedmx}; SQ combines them with shared bit appending and width reduction to target conventional integer operands.

\subsection{SHI four-bit products and grouped corrections}
\label{sec:shi}

SQ reconstructs an integer operand before multiplication. For some low-bit formats, that operand requires INT8 even though the stored code has four bits or fewer. We developed Shifter Integer (SHI) as a variant of SQ designed to use native INT4 arithmetic. When the reconstructed weight fits in INT4, it can be used directly; otherwise, SHI evaluates a low-bit base and a shared group correction through separate INT4 products. Ignoring channel scales for now, the latter representation can be written as
\begin{equation}
    w_{j,k}=\rho q_{j,k}+r_{j,g(k)},
    \label{eq:shiweight}
\end{equation}
Here $q$ is the base operand for an individual weight, $r$ is a signed correction shared by its group, and $\rho$ sets the relative size of the two parts. The operands use a balanced representation of the stored fields; the exact conversion and any clipping are part of the numerical profile. SHI2--SHI4 use INT2 through INT4 base codes, so their base operands fit in signed INT4 and can use INT4 arithmetic. A five-bit or wider base would require a wider operand, such as INT8, and would not retain this advantage. For example, SHI4 G8 uses a four-bit base per weight and one four-bit correction per eight weights. These fields occupy $4+4/8=4.5$ bits per weight, before scales and other metadata.

Like other groupwise formats, SHI spends extra bits to improve the representation of each group. NVFP4 stores an eight-bit scale per 16 four-bit values, giving 4.5 bits per value before global scales and padding; MXFP4 stores an eight-bit scale per 32 values, giving 4.25 bits per value~\citep{nvfp4,mxspec}. When both weights and activations use these formats, their effective rates are therefore W4.5A4.5 and W4.25A4.25, respectively. On hardware with native block-scaled FP4 instructions, the scales are applied within the matrix operation, allowing these formats to use native FP4 matrix throughput. SHI instead evaluates its group corrections through integer products.

The following SQ field structures illustrate how SHI preserves four-bit arithmetic:
\begin{itemize}
    \item \textbf{SHI2: SQ2-G4A1S1.} The expanded and shifted weight fits in a signed INT4 operand. It can therefore be reconstructed within the kernel and multiplied directly by an INT4 activation in one product. Because the activation has more bits than the two-bit base code, we primarily use the single-product path without a separate activation correction.
    \item \textbf{SHI3: SQ3-G8A3S1.} The three-bit base and one-bit shift allow a four-bit main operand; the three append bits and shift allow a four-bit grouped correction operand under the selected reconstruction and range handling. We use two products to evaluate these weight components. The four-bit activation has more bits than the base code, so we primarily use the two-product path without an activation correction.
    \item \textbf{SHI4: SQ4-G8A4S0.} With no shift selector, the four-bit base and four-bit shared append field provide the main and grouped weight components; a balanced representation maps them to signed operands. Because the activation also uses INT4, we use a corresponding SQ4-G8A4S0-style activation representation and a third product to include its grouped correction.
\end{itemize}
For SHI4 G8, we refer to the base-only, weight-corrected, and weight- and activation-corrected computations as W4A4, W4.5A4, and W4.5A4.5, respectively. The fractional bit counts describe the shared four-bit fields averaged over eight values; they do not change the INT4 operand width. The same product structures also apply to other SQ formats, whose effective bit counts may differ.

For SHI2, the single-product path uses the complete reconstructed INT4 weight. For a representation split into a base and correction, the base-only expression below illustrates the effect of omitting the correction. The weight- and activation-correction equations use the unsaturated representation in Equation~\eqref{eq:shiweight}; clipping a combined INT8 operand changes values at clipped entries.

\subsubsection{Main product (W4A4)}

The simplest option multiplies the weight base by a four-bit approximation of the activations. Let $\boldsymbol{\xi}$ be an input row after rotation and input-channel scaling, and approximate each activation by $\xi_k\simeq s a_k$, where $a_k$ is a four-bit code and $s$ is its scale. Using only the weight base gives
\begin{equation}
    \widehat y^{\mathrm{base}}_j
      =c_{o,j}\rho s\sum_k q_{j,k}a_k.
\end{equation}
This is one INT4 product across all $K$ input channels. It is the least expensive option, but it leaves out the stored weight correction.

\subsubsection{Grouped weight correction (W4.5A4)}

The weight-corrected path adds the shared weight correction. Because every weight in a group shares the same correction, we can first sum the activations in that group and then multiply the sum by the correction:
\begin{equation}
    \sum_k(\rho q_{j,k}+r_{j,g(k)})\xi_k
    =\rho\sum_k q_{j,k}\xi_k
      +\sum_g r_{j,g}\sum_{k\in g}\xi_k.
    \label{eq:2exact}
\end{equation}
The first product still covers all $K$ weights. The correction product needs only one value per group, so it has length $K/y$: for groups of eight, it is one eighth as long as the main product.

To run the correction product in INT4, we sum the activation codes, $h_g=\sum_{k\in g}a_k$, and quantize the scaled sums as $s h_g\simeq t b_g$. Here $b_g$ is a four-bit code and $t$ is a per-token scale. The two products compute
\begin{equation}
    \widehat y^{\mathrm{weight}}_j
      =c_{o,j}\left[
         \rho s\sum_k q_{j,k}a_k
         +t\sum_g r_{j,g}b_g
      \right].
    \label{eq:2exec}
\end{equation}
The second product approximates the contribution of the weight correction. The group sums can exceed the INT4 range, so quantizing them introduces error. The activations themselves still use only their base four-bit approximation.

\subsubsection{Grouped activation correction (W4.5A4.5)}

The activation-corrected path adds a correction shared by each activation group:
\begin{equation}
    \xi_k\simeq s(\tau a_k+c_{g(k)}),
    \label{eq:activation}
\end{equation}
Here $a$ and $c$ are four-bit codes and $\tau$ sets their relative size. To calculate the corrections efficiently, define the sum of the activation bases and the sum of the reconstructed weights in each group:
\begin{equation}
    S_{a,g}=\sum_{k\in g}a_k,
    \qquad
    S_{w,j,g}=\rho\sum_{k\in g}q_{j,k}+y r_{j,g}.
\end{equation}
The product of the weight and activation representations can then be written as
\begin{equation}
    y_j\simeq c_{o,j}s\left[
       \rho\tau\sum_k q_{j,k}a_k
       +\tau\sum_g r_{j,g}S_{a,g}
       +\sum_g S_{w,j,g}c_g
    \right].
    \label{eq:3exact}
\end{equation}
The first term is the main product, the second accounts for the weight correction, and the third accounts for the activation correction. The weight sum includes $y r$, so the third term also accounts for the interaction between the two corrections; no fourth product is needed.

The group sums can be wider than INT4, so we quantize them before the two correction products:
\begin{equation}
    s\tau S_{a,g}\simeq t b_g,
    \qquad
    S_{w,j,g}\simeq u_j d_{j,g},
\end{equation}
Here $b$ and $d$ are four-bit codes, $t$ is a per-token scale, and $u_j$ is a scale fitted per output row. The executed three-product approximation is
\begin{equation}
    \widehat y^{\mathrm{weight+activation}}_j=c_{o,j}\left[
       \rho\tau s\sum_k q_{j,k}a_k
       +t\sum_g r_{j,g}b_g
       +s u_j\sum_g d_{j,g}c_g
    \right].
    \label{eq:3exec}
\end{equation}
In the evaluated SHI4 construction, $\rho=\tau=16$ and all three products use four-bit operands. Quantizing the group sums introduces additional error, so the executed result in Equation~\eqref{eq:3exec} approximates the expansion in Equation~\eqref{eq:3exact}.

\begin{table}[H]
\centering
\begin{tabularx}{\linewidth}{@{}l L c c c@{}}
\toprule
Profile & Information evaluated & Reduction lengths & Work at G8 & Work at G16\\
\midrule
One product & Base weight and activation & $K$ & $1.000K$ & $1.000K$\\
Two products & Add grouped weight correction & $K+K/y$ & $1.125K$ & $1.0625K$\\
Three products & Also correct activations, including the cross term & $K+2K/y$ & $1.250K$ & $1.125K$\\
\bottomrule
\end{tabularx}
\caption{Logical contraction work per output, where $K$ is the work of the W4A4 main product.}
\label{tab:profiles}
\end{table}

SHI is designed to use the structure of the quantized weights during matrix multiplication, avoiding a full-length INT8 product when INT4 products are available. As a throughput-normalized comparison, if INT4 has twice the arithmetic throughput of INT8, an INT8 product of length $K$ corresponds to $2K$ of INT4 work. Table~\ref{tab:profiles} counts only the INT4 contractions, not preparation or data movement. For groups of eight, the two- and three-product paths require $1.125K$ and $1.25K$ of work; with groups of sixteen, these fall to $1.0625K$ and $1.125K$. Larger groups reduce correction work but may affect quantization accuracy. Multiple products can be fused into one kernel, so the number of products need not equal the number of launches.

A prepared weight-sum plane $d$ can increase resident memory. If four-bit $q$, $r$ and $d$ are all retained, they require $4+8/y$ logical code bits per weight, excluding scales and padding. For G8, the serialized base and correction occupy 4.5 bits per weight, while this prepared three-plane representation occupies 5.0. The derived plane is included in resident-memory accounting when retained, even if it is generated at load time.

Details of activation quantization and numerical profiles are given in Appendix~\ref{app:activation}.

\section{Projection-level evaluation}
\label{sec:projection}

We first isolate one MLP projection to measure the accuracy and latency of
each quantization path.

\subsection{Benchmark protocol}

We isolate the layer-31 MLP down projection of Qwen3.8-27B
($K=17{,}408$, $N=5{,}120$) on an RTX 5090. All formats are fitted to the
same BF16 weights using unweighted squared reconstruction error. We measure
weight error by the relative Frobenius norm and output error by the mean
relative $L_2$ error on 255 activation vectors per WikiText-2 context. We set
the residual to zero when measuring output error and round the output to
BF16.

We time each projection at 4,096 tokens. For cold-cache timing, llama.cpp starts with resident FP32 inputs and a
ready FP32 residual and writes FP32 output. The Shifter and floating-point
CUDA paths use BF16 inputs, residuals and outputs.
The timed interval includes activation preparation, matrix multiplication, scaling,
residual addition, backend conversions, host dispatch and synchronization.
The tables report median latency and use llama.cpp Q2\_K and Q4\_K as the
respective throughput baselines. Q2\_K also serves as the reference format
in the full-checkpoint comparison (Table~\ref{tab:full-model-2bit}). Packed bits per weight (bpw) count weight codes and group
fields. Shifter also stores channel scales and H512 signs. The llama.cpp
formats serve as software baselines~\citep{llamacpp}.

\subsection{Two-bit projection results}

\begin{table}[H]
\centering
\footnotesize
\setlength{\tabcolsep}{3pt}
\begin{tabularx}{\linewidth}{@{}L l r r r r r@{}}
\toprule
Weight / engine & Activation & \shortstack{Packed\\bpw} & \shortstack{Weight\\error} &
\shortstack{Output\\error} & \shortstack{4,096-token\\latency} & \shortstack{Speedup\\vs. Q2\_K}\\
\midrule
\ProjectionTwoBitRows
\bottomrule
\end{tabularx}
\caption{Two-bit projection error and latency. Speedup uses the llama.cpp Q2\_K baseline.}
\label{tab:projection-2bit}
\end{table}

We measure output error on five WikiText-2 test contexts. SQ2 and SHI2
share one SQ2-G4A1S1 weight fit. SQ2 uses INT8 activations, and SHI2 uses the
single-product W4A4 path described in Section~\ref{sec:shi}.
The custom Q2\_0 comparison format groups 32 weights per FP16 scale,
for 2.500 packed bpw.

At the same 2.500 packed bits per weight, G4 SQ2 with INT8 activations
reduces output error from 30.680\% to 22.971\% and achieves $1.68\times$ the
throughput of G32 Q2\_0. Against Q2\_K, SQ2 reduces output error from 24.152\%
to 22.971\% and achieves $2.77\times$ the throughput while using 2.500 packed
bits per weight versus 2.625.

SQ2 and SHI2 use the same fitted weights in this experiment, so the comparison
shows what changes when we switch from INT8 activations to the single-product
INT4 path. SHI2 delivers $1.65\times$ the projection throughput, reducing
latency from 1,504.74 to 914.34\,\textmu s. Output error rises from 22.971\%
to 27.924\%.

\subsection{Four-bit projection results}

\begin{table}[H]
\centering
\footnotesize
\setlength{\tabcolsep}{3pt}
\begin{tabularx}{\linewidth}{@{}L l r r r r r@{}}
\toprule
Weight / engine & Activation & \shortstack{Packed\\bpw} & \shortstack{Weight\\error} &
\shortstack{Output\\error} & \shortstack{4,096-token\\latency} & \shortstack{Speedup\\vs. Q4\_K}\\
\midrule
\ProjectionFourBitRows
\bottomrule
\end{tabularx}
\caption{Four-bit projection error and latency. Speedup uses the Q4\_K
baseline.}
\label{tab:projection-4bit}
\end{table}

Each pair of SHI4, NVFP4 or MXFP4 rows shares one fitted weight tensor.
The Shifter paths use H512, while the llama.cpp and floating-point paths
use the original basis. The SHI4 activation row uses Shifter Engine's native
three-product down projection. We use the first captured context to
calibrate a fixed tensor-wide NVFP4 activation scale and two fixed SHI4
clipping fractions. For SHI4, we search five factors each for base
activations and G8 base-code sums. NVFP4 computes its G16 scales online.
We measure output error for all formats on the remaining four contexts.

The llama.cpp CUDA graph prepares Q8\_1 activations from resident FP32 input. FP8
timings include expanding packed weights to FP8 on each call and use a
tuned cuBLASLt schedule. The MXFP4 variant fits one E8M0 scale per 16 weights
for 4.5 packed bpw.

SHI4 with four-bit activations is $2.83\times$ as fast as Q4\_K on this
projection, with 12.236\% output error. SQ4-G8A2S2/INT8 is $1.94\times$
as fast as Q4\_K, with 6.804\% output error versus 6.688\% for Q4\_K.

\section{Full-model evaluation}
\label{sec:full-model-quality}

We next evaluate complete checkpoints to measure whether the projection-level results carry over to model quality and prefill. We compare two-, four-, and eight-bit formats separately.

\subsection{WikiText-2 protocol}

We use WikiText-2~\citep{wikitext} because llama.cpp, a widely used local inference engine,
provides a standard procedure for comparing quantized models with their
higher-precision references. Its evaluation reports KL divergence, top-1
agreement and perplexity, metrics also used in community quantization
comparisons~\citep{llamacppperplexity,unslothdynamic}.

Each model is scored on 147,900 next-token targets from 580 WikiText-2 test
contexts using llama.cpp's probability and KL implementation. Each checkpoint
uses the corresponding original BF16 model as the reference. The checkpoint names identify
their main quantization format within a mixed-precision model.

KL divergence is our primary quality metric because it measures how much
quantization changes the model's next-token probability distribution relative
to the BF16 reference. Top-1 agreement is the fraction of positions where the quantized and BF16
models predict the same top token. Perplexity (PPL) measures prediction on
the test text.

We evaluate weight-only fits without calibration and also compare released
checkpoints. When the provider's calibration data is available, we use it to
fit the corresponding Shifter checkpoint.

\subsection{Two-bit checkpoints}
\label{sec:full-model-2bit}

\subsubsection{Weight-only fitting}

\begin{table}[H]
\centering
\small
\setlength{\tabcolsep}{5pt}
\begin{tabularx}{0.80\linewidth}{@{}L r r r@{}}
\toprule
Weight-only model & KL $\downarrow$ & Top-1 $\uparrow$ & PPL $\downarrow$ \\
\midrule
Shifter-SQ2 & \textbf{0.20001} & \textbf{81.6633\%} & 7.91754 \\
Shifter-SHI2 & 0.21210 & 80.9838\% & \textbf{7.89657} \\
Q2\_K recipe (local refit) & 0.27849 & 78.2596\% & 8.58626 \\
\bottomrule
\end{tabularx}
\caption{Full-checkpoint WikiText-2 quality for weight-only fits.}
\label{tab:full-model-2bit-weight-only}
\end{table}

We evaluate the weight-only fits on the same contexts and targets.
The Q2\_K comparison refits Bartowski's mixed-precision recipe without
calibration.

The two-bit projection results show that SQ2 can reduce output error at the
same weight storage as G32 Q2\_0. At full-model scale, the weight-only fits
show that the quality advantage extends to next-token predictions without
calibration. SQ2 and SHI2 reduce WikiText-2 KL by 28.2\% and 23.8\%,
respectively, relative to the Q2\_K refit, and both improve top-1 agreement
and perplexity.

\subsubsection{Released checkpoint comparison}

\begin{table}[H]
\centering
\small
\setlength{\tabcolsep}{5pt}
\begin{tabularx}{\linewidth}{@{}L r r r r@{}}
\toprule
Model & KL $\downarrow$ & Top-1 $\uparrow$ & PPL $\downarrow$ &
\shortstack{4,096-token prefill\\(tok/s) $\uparrow$} \\
\midrule
Shifter-SQ2 & \textbf{0.15322} & \textbf{84.0237\%} & \textbf{7.57241} & 8,378.4 ($2.71\times$) \\
\ShifterSHITwoRow
Bartowski Q2\_K & 0.16627 & 83.3110\% & 7.70266 & 3,090.7 ($1.00\times$) \\
\bottomrule
\end{tabularx}
\caption{Full-checkpoint quality and prefill throughput. Speedups are
relative to Bartowski Q2\_K.}
\label{tab:full-model-2bit}
\end{table}

We compare Shifter-SQ2 and Shifter-SHI2 for Qwen3.8-27B with Bartowski's
released Q2\_K checkpoint in llama.cpp. Table~\ref{tab:full-model-2bit}
pairs WikiText-2 quality with 4,096-token prefill throughput.

Both Shifter checkpoints use INT8 activations for quality evaluation.
The quality evaluation uses llama.cpp's probability helper for the same
contexts and targets. For prefill, Shifter-SQ2 uses INT8 activations and
Shifter-SHI2 uses four-bit activations. Each prefill
run uses a single 4,096-token chunk, BF16 KV and fresh prompt state. We
report the median of three runs after two warmups.

Against Bartowski's released Q2\_K checkpoint, SQ2 achieves $2.71\times$ the
prefill throughput and has the lowest KL and perplexity and the highest top-1
agreement in this comparison. SHI2 reaches \ShifterSHITwoPrefill{} tokens/s
with four-bit activations, or $\ShifterSHITwoSpeedup\times$ the Q2\_K prefill
throughput. With INT8 activations for the quality evaluation, SHI2 also
improves all three WikiText-2 metrics over Q2\_K.

\subsection{Four-bit checkpoint pairs}
\label{sec:full-model-4bit}

\begin{table}[H]
\centering
\small
\setlength{\tabcolsep}{5pt}
\begin{tabularx}{\linewidth}{@{}L r r r@{}}
\toprule
Model & KL $\downarrow$ & Top-1 $\uparrow$ & PPL $\downarrow$ \\
\midrule
\QualityRows
\bottomrule
\end{tabularx}
\caption{Full-checkpoint WikiText-2 quality. Bold marks the better value within each pair.}
\label{tab:full-model-4bit}
\end{table}

For Qwen3.8-27B we compare SHI4 against NVFP4 and SQ4 against groupwise INT.
For Qwen3.6-35B-A3B we compare SQ4 against groupwise INT. All six runs use
BF16 KV.

The SQ4 checkpoints use INT8 activation arithmetic on their SQ4 projections.
The SHI4 quality evaluation uses the MACaQ-calibrated GGUF~\citep{macaq} with A16
activations on the balanced MLPs and outlier protection on eligible ancillary
INT8 projections. NInfer uses BF16 activations for NVFP4 projections.

SHI4 reduces WikiText-2 KL by 26.8\% relative to NVFP4, with higher top-1
agreement and lower perplexity.
Both SQ4 checkpoints improve all three metrics against their groupwise INT
baselines.

\subsection{Eight-bit checkpoint comparison}
\label{sec:full-model-8bit}

\begin{table}[H]
\centering
\small
\setlength{\tabcolsep}{5pt}
\begin{tabularx}{\linewidth}{@{}L r r r r@{}}
\toprule
Runtime & KL $\downarrow$ & Top-1 $\uparrow$ & PPL $\downarrow$ &
\shortstack{4,096-token prefill\\(tok/s) $\uparrow$} \\
\midrule
\WEightAEightRows
\bottomrule
\end{tabularx}
\caption{Eight-bit full-model quality and batch-one prefill.
Prefill uses a single 4,096-token chunk, fresh state, and one greedy next-token
result. Rates are medians of three measurements after two warmups.}
\label{tab:full-model-8bit}
\end{table}

At the eight-bit end, we compare Shifter Engine's SQ8 checkpoint with
vLLM's online FP8 W8A8 on Qwen3.8-27B~\citep{vllm,vllmonline}. Shifter fits per-output-channel
INT8 weights after the signed Hadamard rotation and uses flat per-token INT8
activation scales. vLLM quantizes the BF16 checkpoint to per-tensor E4M3
weights at load time and uses dynamically scaled FP8 activations through
CUTLASS. Both use BF16 KV and the same BF16 WikiText-2 reference, with
CUDA Graphs, prefix reuse, and speculative decoding disabled for timing.

Shifter prefills \WEightAEightSpeedGain\% faster, reduces KL by
\WEightAEightKLGain\%, and improves top-1 agreement by
\WEightAEightAgreementGain{} percentage points. The PPL values are nearly
identical.

\section{Four-bit serving evaluation}
\label{sec:serving}

We compare Shifter Engine with NInfer, a community-developed local inference engine specialized for selected Qwen models~\citep{ninfer}. NInfer has also inspired forks targeting other GPUs and deployment environments~\citep{ninferfork}. Its specialization and published performance make it a strong community baseline for these models. We evaluate three four-bit model pairs on single-request prefill and structured decode, followed by concurrent decode.

\subsection{Deployment configurations}

The serving comparisons use the same three model and format pairs as
Table~\ref{tab:full-model-4bit}. For serving,
Shifter uses the MACaQ-calibrated SHI4 artifact~\citep{macaq} with statically clipped
four-bit MLP activations and disables outlier protection on ancillary INT8
projections.

The RTX 5090 Shifter measurements use Linux with a 450\,W power limit,
driver 595.71.05, CUDA 13.2, and PyTorch 2.9.1+cu128. Shifter enables
short-batch MoE expert-weight reuse. The serving tables use NInfer's
published rates for revision \texttt{a754fcf3}. Shifter uses normalized H256
on attention Q/K/V and
INT8 KV with one FP16 scale per head vector. NInfer uses FP8 E4M3 G128 KV
for 27B and G64 INT8 KV for 35B.

\subsection{Single-request prefill and structured decode}

\begin{table}[H]
\centering
\small
\setlength{\tabcolsep}{4pt}
\begin{tabularx}{\linewidth}{@{}L r r r@{}}
\toprule
Model and comparison & \shortstack{Prefill\\7.7K} & \shortstack{Prefill\\260K} & \shortstack{Structured\\MTP3}\\
\midrule
\SingleServingRows
\bottomrule
\end{tabularx}
\caption{Single-request serving throughput in tokens/s. Shifter structured MTP3 reports mean $\pm$ sample standard deviation over three requests for each 27B checkpoint and 15 for 35B. NInfer structured MTP3 values are means. Bold marks the higher measured throughput
within each pair.}
\label{tab:serving-single}
\end{table}

For prefill, we use NInfer's 7,680- and 260,096-token needle-retrieval
fixtures and disable speculation. We use its JSONL, CSV and SQL fixtures for
structured decode with three-token multi-token prediction (MTP3) and the
optimized proposal head~\citep{specdecode,mtp}. Shifter enables CUDA graphs, uses 4,096-token chunks
for short prefill and 1,024-token chunks for long prefill and decode, and
disables both prefix reuse and thinking. Sampling uses temperature
0.6, top-$p$ 0.95, top-$k$ 20, minimum probability 0, presence penalty 1 and
frequency penalty 0. Structured MTP3 uses a 4,096-token output cap per request. Each model is prepared
once per phase sweep, with fresh sequence state for every request.

Table~\ref{tab:serving-single} divides input tokens by server prefill-phase
time and averages the first three fixed seeds for the short prompt. The
extreme prompt uses the first fixed seed. Shifter generates one output token in each
prefill run. We compute each
request's decode rate by dividing the number of output tokens after the
first by the recorded decode time.
Shifter uses one seed per fixture for 27B and five per fixture for 35B.
We retain actual termination and completion lengths and exclude loading
and warmup.

Relative to groupwise INT, SQ4 reaches $2.47\times$ and $1.43\times$
the short- and extreme-prefill throughput on 27B, and $1.81\times$ and
$1.67\times$ on 35B. On structured decode, the 35B means are 784.0 tokens/s
for Shifter and 779.6 tokens/s for NInfer.

\subsection{Concurrent MTP3 decode}

\begin{table}[H]
\centering
\footnotesize
\setlength{\tabcolsep}{3pt}
\begin{tabularx}{\linewidth}{@{}L r r r r@{}}
\toprule
Model and comparison & $C=1$ & $C=2$ & $C=4$ & $C=8$\\
\midrule
\ConcurrentServingRows
\bottomrule
\end{tabularx}
\caption{Concurrent MTP3 decode. Each cell reports steady throughput in
tokens/s followed by full-wave acceptance. Bold marks the higher measured
throughput within each pair.}
\label{tab:serving-concurrent}
\end{table}

For each concurrency $C\in\{1,2,4,8\}$, we issue one request per
sequence using NInfer's
\texttt{long\_\allowbreak decode\_\allowbreak aime26\_\allowbreak 15} fixture. Requests use distinct fixed seeds,
thinking enabled, an 8,192-token output budget and a 16,384-token context
limit. CUDA graphs, MTP3, the proposal head, sampling and prefill chunk
size follow the single-request setup. Requests can end at EOS. The
Shifter benchmark reuses one prepared deployment for compatible waves.
For 27B SQ4 it pins tiled weights through eight verifier columns, and for
27B SHI4 it pins the tiled gate/up and down bindings through sixteen.
Wider waves and 35B SQ4 use row-major weights. Structured decode uses the
same narrow layout choices.

Table~\ref{tab:serving-concurrent} counts committed decode tokens in
complete one-second intervals where all $C$ sequences are decode-ready,
every actual decode batch has size $C$, and no prefill or recovery occurs.
Acceptance divides
accepted tokens by drafted tokens over the complete request wave. Each cell
represents one wave.

At concurrency eight, Shifter delivers $0.88\times$, $1.31\times$ and
$1.25\times$ NInfer's throughput for 27B SHI4, 27B SQ4 and
35B SQ4, respectively.

\section{Cross-GPU performance scaling}
\label{sec:cross-gpu}

Both SQ4 checkpoints run on four GPUs spanning Ampere, Ada, and Blackwell.
On both Ampere GPUs, 35B-A3B SQ4 exceeds 10,800 tokens/s in prefill and
330 tokens/s in structured MTP3 decode. Across the four GPUs, prefill
throughput tracks peak dense INT8 throughput, with $R^2=0.987$--$0.992$,
while structured decode throughput tracks memory bandwidth, with
$R^2=0.939$--$0.970$~\citep{amperegpu,blackwellgpu}.
Figure~\ref{fig:sq4-cross-gpu} shows the fits, and Table~\ref{tab:sq4-cross-gpu}
in the appendix reports the measured rates.

\begin{figure}[H]
\centering
\includegraphics[width=\linewidth]{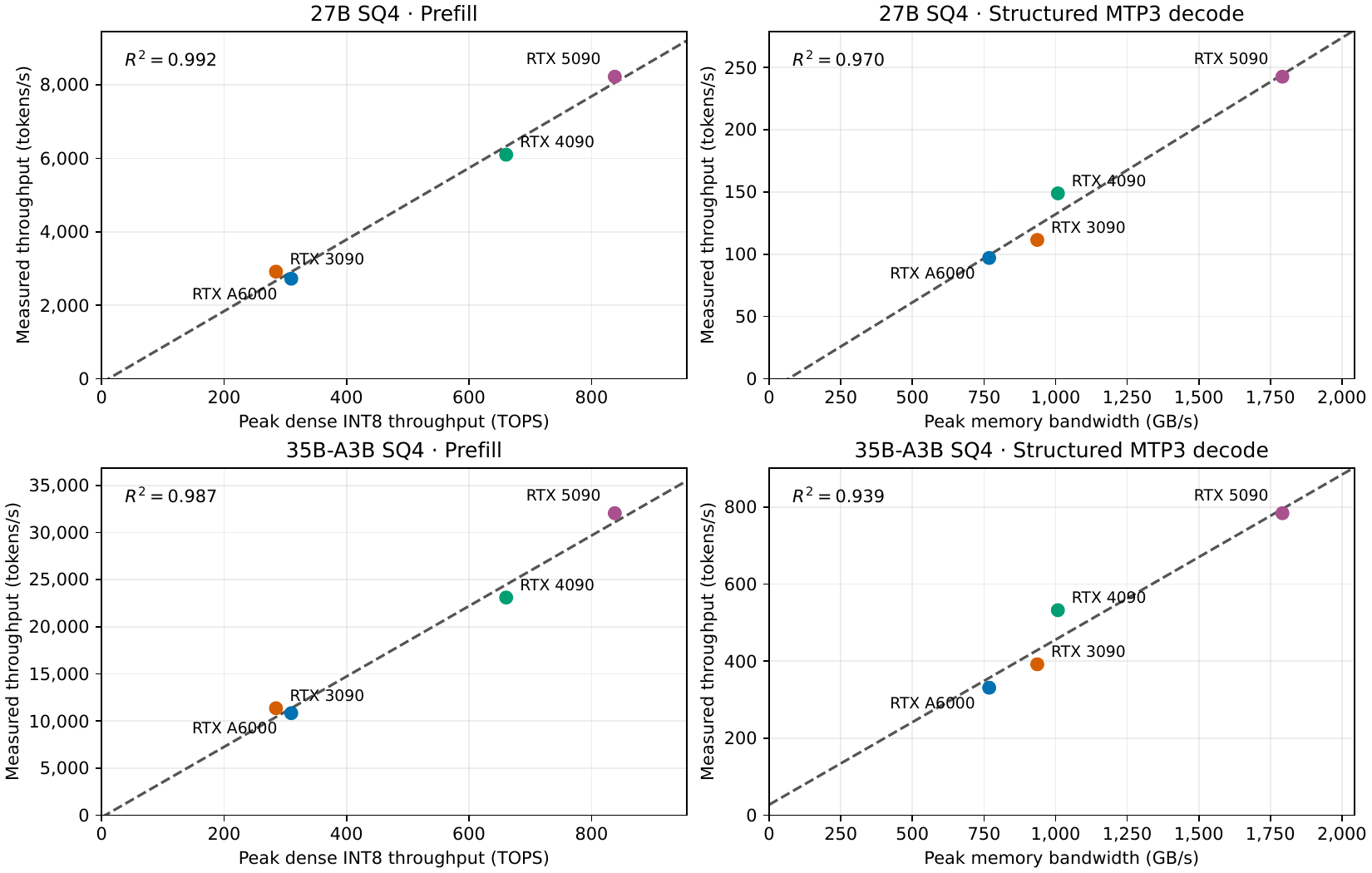}
\caption{SQ4 throughput across four GPUs for 7,680-token prefill and
single-request structured MTP3 decode. Dashed lines show least-squares fits
with a free intercept. Prefill chunk sizes, context allocations, and seed
counts vary across deployments.}
\label{fig:sq4-cross-gpu}
\end{figure}

\section{Conclusion}

We introduced Shifter Quantization (SQ) and Shifter Integer (SHI) for
low-bit inference with conventional integer matrix operations. SQ stores
compact weight codes with shared group information and reconstructs integer
operands through bitwise operations. This places local adaptation in the
operand rather than requiring a floating-point group scale inside the matrix
reduction. SHI builds on this representation to use four-bit products,
evaluating grouped weight and activation corrections through shorter integer
products when needed. Both approaches retain compact weights while allowing
the main product to accumulate across the full reduction dimension. This
structure aligns the computation with hardware-accelerated integer matrix
instructions, helping translate low-bit storage into faster inference.

In the weight-only two-bit comparison, SQ2 and SHI2 reduce WikiText-2 KL by 28.2\% and 23.8\%,
respectively, relative to a Q2\_K recipe refit without calibration. SQ and
SHI can assign shifts or corrections to small groups without introducing a
floating-point scale for each group inside the main reduction. Smaller groups let SQ and SHI
adapt to local weight variation. SQ2 also
achieves $2.77\times$ the Q2\_K throughput on the layer-31 down projection with
lower output error. Against the released Q2\_K checkpoint, SQ2 and SHI2
reach $2.71\times$ and $3.77\times$ the 4,096-token prefill throughput,
respectively.

SQ4 reduces WikiText-2 KL by 24--40\% against NInfer's groupwise integer
checkpoints and achieves $1.43$--$2.47\times$ their NInfer prefill
throughput across the two Qwen models and prompt lengths. At concurrency
eight, SQ4 reaches $1.31\times$ and $1.25\times$ the decode throughput of the
corresponding NInfer checkpoints for the 27B and 35B models, respectively.
At eight bits, SQ8 reduces WikiText-2 KL by \WEightAEightKLGain\% and
improves batch-one prefill throughput by \WEightAEightSpeedGain\% over
online FP8.

SQ and SHI show that small groups can preserve weight accuracy while still
letting the GPU use integer MMA across the full dot product. This combination
makes it possible to improve both model quality and prefill speed with compact
weights.

\bibliographystyle{abbrvnat}
\bibliography{shifter-quantization}

\appendix
\section{SQ worked example}
\label{app:sq-example}
Consider four transformed and channel-scaled weights
$(-0.28,-0.12,0.09,0.31)$. To illustrate SQ2-G4A1S1, choose the base
codes $q=(-2,-1,0,1)$, a shared append bit $a_g=1$, and a shift selector
$r_g=1$. Let the fixed alignment be $\ell=2$, the final reduction be
$\delta=1$, and the output operand width be $t=8$. The three operations give
\begin{align*}
    z &= 2q+1=(-3,-1,1,3),\\
    v &= (4z)\mathbin{\gg}1=(-6,-2,2,6),\\
    P &= \clip_8(v\mathbin{\gg}1)=(-3,-1,1,3).
\end{align*}
All four values fit in INT8. With a row output scale of $0.1$, they
reconstruct to $(-0.3,-0.1,0.1,0.3)$. Setting the append bit to zero
with the same base codes and shifts gives $(-0.4,-0.2,0,0.2)$.
The shared bit therefore moves all four reconstructed weights by $0.1$
while costing only one bit for the group.

\section{Activation quantization and numerical profiles}
\label{app:activation}

\subsection{A4 clipping and grouped correction}
For ordinary A4 quantization, choose a per-token step $S$ and form $a=\clip(\round(x/S))$. The dynamic selection used in the evaluated implementation minimizes activation reconstruction error over clipping factors
\begin{equation}
    \alpha\in\{0.5,\,0.625,\,0.75,\,0.875,\,1\},
    \qquad
    S_\alpha=\max(\alpha\lVert x\rVert_\infty/7,\epsilon).
\end{equation}
Codes use nearest-even rounding and the specified clipping range. The selected scale is shared across the token. A static variant obtains the clipping factor from calibration while retaining token-dependent magnitude estimation. We evaluate static and dynamic clipping separately.

For the activation representation with grouped correction, set $s=S/16$ after ordinary scale selection. At this scale, fit $a$ and $c$ using two shared-correction initializations, 0 and 7. For each initialization, alternate clipped main-code rounding with a clipped, rounded group-mean residual correction, then perform one further update of each. Choose the lower squared-error result independently per group. The main codes preceding the final correction update are retained. The activation-sum conversion has its own declared scale and clipping rule. The evaluated activation preparation rounds the rotated values to BF16 before input-channel scaling; fusing these stages preserves that rounding boundary.

\subsection{A8 and A16 execution}
A8 quantizes activations with per-token scales and executes INT8 products with INT32 accumulation, using the decoded SQ operand $P$ directly. A16 uses the specified BF16 or FP16 arithmetic and converts packed weights on demand or through staging. For small token counts, avoiding activation quantization can offset the cost of wider arithmetic. This path can operate without materializing the complete floating-point weight matrix.

Optional A8 outlier protection excludes selected large activations from the main quantized product and adds their contributions using the original values. The RMS threshold, per-token cap and correction arithmetic determine both its accuracy and cost.

\section{Cross-GPU throughput measurements}
\label{app:cross-gpu}

\begingroup
\setlength{\intextsep}{5pt}
\begin{table}[H]
\centering
\small
\setlength{\abovecaptionskip}{5pt}
\renewcommand{\arraystretch}{0.9}
\begin{tabular}{@{}lrrrr@{}}
\toprule
& \multicolumn{2}{c}{27B SQ4} & \multicolumn{2}{c}{35B-A3B SQ4} \\
\cmidrule(lr){2-3}\cmidrule(lr){4-5}
GPU & Prefill & MTP3 decode & Prefill & MTP3 decode \\
\midrule
RTX A6000 & 2,723.8 & 97.0 & 10,828.8 & 331.0 \\
RTX 3090 & 2,915.7 & 111.5 & 11,354.8 & 391.9 \\
RTX 4090 & 6,098.7 & 148.9 & 23,093.3 & 532.3 \\
RTX 5090 & 8,221.9 & 242.6 & 32,039.0 & 784.0 \\
\bottomrule
\end{tabular}
\caption{SQ4 throughput (tokens/s) for Figure~\ref{fig:sq4-cross-gpu}.
Prefill uses 7,680 tokens. Structured MTP3 uses JSONL, CSV, and SQL,
with five seeds per fixture for 35B-A3B on RTX 5090 and one elsewhere.}
\label{tab:sq4-cross-gpu}
\end{table}
\endgroup

\end{document}

%% file: results/serving-tables.tex
\newcommand{\SingleServingRows}{%
Qwen3.8-27B SHI4 / Shifter & 10,252.3 & 3,550.0 & 212.1$\pm$24.0 \\
Qwen3.8-27B NVFP4 / NInfer & \textbf{12,819.1} & \textbf{4,016.4} & \textbf{231.7} \\
\textit{Shifter / NInfer} & 0.80$\times$ & 0.88$\times$ & 0.92$\times$ \\
\midrule
Qwen3.8-27B SQ4 / Shifter & \textbf{8,221.9} & \textbf{3,055.2} & \textbf{242.6$\pm$16.6} \\
Qwen3.8-27B groupwise INT / NInfer & 3,331.9 & 2,139.4 & 214.7 \\
\textit{Shifter / NInfer} & 2.47$\times$ & 1.43$\times$ & 1.13$\times$ \\
\midrule
Qwen3.6-35B-A3B SQ4 / Shifter & \textbf{32,039.0} & \textbf{8,738.0} & \textbf{784.0$\pm$37.6} \\
Qwen3.6-35B-A3B groupwise INT / NInfer & 17,705.4 & 5,247.0 & 779.6 \\
\textit{Shifter / NInfer} & 1.81$\times$ & 1.67$\times$ & 1.01$\times$ \\
}
\newcommand{\ConcurrentServingRows}{%
Qwen3.8-27B SHI4 / Shifter & 144.4 / 46.8\% & 276.6 / 46.5\% & 503.4 / 47.6\% & 809.5 / 46.0\% \\
Qwen3.8-27B NVFP4 / NInfer & \textbf{147.7} / 46.2\% & \textbf{291.0} / 48.7\% & \textbf{522.2} / 45.8\% & \textbf{922.4} / 46.1\% \\
\midrule
Qwen3.8-27B SQ4 / Shifter & \textbf{156.9} / 46.1\% & \textbf{283.6} / 45.8\% & \textbf{452.3} / 47.1\% & \textbf{765.5} / 46.7\% \\
Qwen3.8-27B groupwise INT / NInfer & 136.5 / 44.4\% & 253.3 / 45.2\% & 398.1 / 46.1\% & 582.4 / 46.4\% \\
\midrule
Qwen3.6-35B-A3B SQ4 / Shifter & \textbf{654.3} / 68.3\% & \textbf{1,004.5} / 69.0\% & \textbf{1,237.2} / 69.4\% & \textbf{1,721.5} / 68.8\% \\
Qwen3.6-35B-A3B groupwise INT / NInfer & 642.5 / 68.6\% & 907.2 / 66.3\% & 1,213.5 / 69.6\% & 1,380.7 / 68.0\% \\
}

%% file: results/model-quality.tex
\newcommand{\QualityRows}{%
Qwen3.8-27B SHI4 (A16, protected) / Shifter & \textbf{0.01373} & \textbf{94.9824\%} & \textbf{6.94277} \\
Qwen3.8-27B NVFP4 (BF16) / NInfer & 0.01875 & 93.9249\% & 7.04771 \\
\midrule
Qwen3.8-27B SQ4 / Shifter & \textbf{0.01970} & \textbf{93.9108\%} & \textbf{6.99283} \\
Qwen3.8-27B groupwise INT / NInfer & 0.03256 & 92.1400\% & 7.14469 \\
\midrule
Qwen3.6-35B-A3B SQ4 / Shifter & \textbf{0.01828} & \textbf{94.2042\%} & \textbf{6.75563} \\
Qwen3.6-35B-A3B groupwise INT / NInfer & 0.02400 & 93.3529\% & 6.79117 \\
}

%% file: results/sq8-evaluation.tex
\newcommand{\WEightAEightRows}{%
vLLM FP8 W8A8 & 0.00774 & 96.0297\% & \textbf{6.96035} & 7,226.8 \\
SQ8 / Shifter & \textbf{0.00211} & \textbf{98.0439\%} & 6.96039 & \textbf{9,206.3} \\
}
\newcommand{\WEightAEightSpeedGain}{27.4}
\newcommand{\WEightAEightKLGain}{72.7}
\newcommand{\WEightAEightAgreementGain}{2.01}

%% file: results/shi2-evaluation.tex
\newcommand{\ShifterSHITwoRow}{%
Shifter-SHI2 & 0.15537 & 83.8316\% & 7.58638 & \textbf{11,662.4 ($3.77\times$)} \\
}
\newcommand{\ShifterSHITwoPrefill}{11,662.4}
\newcommand{\ShifterSHITwoSpeedup}{3.77}

%% file: projection/2bit/benchmark-rows.tex
\newcommand{\ProjectionTwoBitRows}{%
Q2\_0 (G32) / llama.cpp & Q8\_1 & 2.500 & 33.574\% & 30.680\% & 2,521.51\,\textmu s & 1.65\ensuremath{\times} \\
Q2\_K / llama.cpp & Q8\_1 & 2.625 & 26.510\% & 24.152\% & 4,162.47\,\textmu s & 1.00\ensuremath{\times} \\
SQ2-G4A1S1 / Shifter (SQ2) & INT8 & 2.500 & \textbf{25.281\%} & \textbf{22.971\%} & 1,504.74\,\textmu s & 2.77\ensuremath{\times} \\
SQ2-G4A1S1 / Shifter (SHI2) & INT4 & 2.500 & \textbf{25.281\%} & 27.924\% & \textbf{914.34\,\textmu s} & \textbf{4.55\ensuremath{\times}} \\%
}

%% file: projection/4bit/benchmark-rows.tex
\newcommand{\ProjectionFourBitRows}{%
Q4\_0 / llama.cpp & Q8\_1 & 4.500 & 8.246\% & 7.496\% & 2,616.33\,\textmu s & 1.10\ensuremath{\times} \\
Q4\_K / llama.cpp & Q8\_1 & 4.500 & \textbf{7.124\%} & \textbf{6.688\%} & 2,873.24\,\textmu s & 1.00\ensuremath{\times} \\
SQ4-G8A2S2 / Shifter & INT8 & 4.500 & 7.457\% & 6.804\% & 1,481.26\,\textmu s & 1.94\ensuremath{\times} \\
SHI4 (G8A4S0) / Shifter & INT8 & 4.500 & 8.632\% & 7.841\% & 1,490.18\,\textmu s & 1.93\ensuremath{\times} \\
 & SHI4 & 4.500 & 8.632\% & 12.236\% & 1,014.05\,\textmu s & 2.83\ensuremath{\times} \\
NVFP4 / CUDA & FP8 & 4.500 & 8.178\% & 7.837\% & 1,472.14\,\textmu s & 1.95\ensuremath{\times} \\
 & NVFP4 & 4.500 & 8.178\% & 10.993\% & 767.71\,\textmu s & 3.74\ensuremath{\times} \\
MXFP4 (E2M1 + E8M0, G16) / CUDA & FP8 & 4.500 & 11.007\% & 10.889\% & 1,473.27\,\textmu s & 1.95\ensuremath{\times} \\
 & NVFP4 & 4.500 & 11.007\% & 12.856\% & \textbf{753.82\,\textmu s} & \textbf{3.81\ensuremath{\times}} \\%
}

%% file: figures/sq4-g8a2s2.tex
\begingroup%
\setlength{\unitlength}{1pt}%
\setlength{\fboxsep}{0pt}%
\definecolor{sqCode}{rgb}{0.88,0.92,0.97}%
\definecolor{sqShared}{rgb}{0.99,0.93,0.80}%
\definecolor{sqOperand}{rgb}{0.87,0.94,0.90}%
\definecolor{sqDropped}{rgb}{0.92,0.92,0.92}%
\definecolor{sqRule}{rgb}{0.63,0.66,0.70}%
\newcommand{\sqbit}[2]{%
  \colorbox{#1}{\makebox[11pt]{\rule[-3pt]{0pt}{13pt}$#2$}}\kern2pt}%
\newcommand{\sqcode}[4]{%
  \begingroup\scriptsize%
  \colorbox{sqCode}{\strut\kern2pt$#1\,#2\,#3\,#4$\kern2pt}%
  \endgroup}%
\newcommand{\sqheading}[1]{\small\bfseries #1}%
\newcommand{\sqnote}[1]{\footnotesize #1}%
\begin{picture}(468,135)
  \put(0,121){\sqheading{Stored codes}}
  \put(302,121){\sqheading{Shared by eight weights}}
  \put(0,102){\sqcode{1}{1}{0}{1}}
  \put(35,102){\sqcode{1}{1}{1}{0}}
  \put(70,102){\sqcode{1}{1}{1}{1}}
  \put(105,102){\sqcode{0}{0}{0}{0}}
  \put(140,102){\sqcode{0}{0}{0}{1}}
  \put(175,102){\sqcode{0}{0}{1}{0}}
  \put(210,102){\sqcode{0}{0}{1}{1}}
  \put(245,102){\sqcode{0}{1}{0}{0}}
  \put(10,88){\sqnote{$q_1$}}
  \put(45,88){\sqnote{$q_2$}}
  \put(80,88){\sqnote{$q_3$}}
  \put(115,88){\sqnote{$q_4$}}
  \put(150,88){\sqnote{$q_5$}}
  \put(185,88){\sqnote{$q_6$}}
  \put(220,88){\sqnote{$q_7$}}
  \put(255,88){\sqnote{$q_8$}}
  \put(302,102){\small\sqbit{sqShared}{1}\sqbit{sqShared}{1}\enspace $a_g=3$}
  \put(386,102){\small\sqbit{sqShared}{1}\sqbit{sqShared}{0}\enspace $s=2$}
  \put(0,78){\color{sqRule}\rule{468pt}{0.4pt}}
  \put(0,60){\sqheading{Base code}}
  \put(97,60){\sqheading{Append shared bits}}
  \put(218,60){\sqheading{Shift left by two}}
  \put(358,60){\sqheading{Truncate to INT8}}
  \put(0,37){\small\sqbit{sqCode}{1}\sqbit{sqCode}{1}\sqbit{sqCode}{0}\sqbit{sqCode}{1}}
  \put(97,37){\small\sqbit{sqCode}{1}\sqbit{sqCode}{1}\sqbit{sqCode}{0}\sqbit{sqCode}{1}%
    \sqbit{sqShared}{1}\sqbit{sqShared}{1}}
  \put(218,37){\small\sqbit{sqCode}{1}\sqbit{sqCode}{1}\sqbit{sqCode}{1}\sqbit{sqCode}{0}%
    \sqbit{sqCode}{1}\sqbit{sqShared}{1}\sqbit{sqShared}{1}\sqbit{sqOperand}{0}\sqbit{sqDropped}{0}}
  \put(358,37){\small\sqbit{sqOperand}{1}\sqbit{sqOperand}{1}\sqbit{sqOperand}{1}\sqbit{sqOperand}{0}%
    \sqbit{sqOperand}{1}\sqbit{sqOperand}{1}\sqbit{sqOperand}{1}\sqbit{sqOperand}{0}}
  \put(60,40){\color{sqRule}\vector(1,0){27}}
  \put(182,40){\color{sqRule}\vector(1,0){27}}
  \put(339,40){\color{sqRule}\vector(1,0){11}}
  \put(0,15){\small $q_1=-3$}
  \put(97,15){\small $z=4q_1+a_g=-9$}
  \put(218,15){\small $z\ll 2=-36$}
  \put(358,15){\small $P=(-36)\gg 1=-18$}
\end{picture}%
\endgroup

%% file: shifter-quantization.bib
@misc{kimik3,
  author = {{Moonshot AI}},
  title = {{Kimi K3 model card}},
  year = {2026},
  url = {https://huggingface.co/moonshotai/Kimi-K3},
}

@misc{qwen38max,
  author = {{Qwen Team}},
  title = {{Qwen 3.8 release announcement}},
  year = {2026},
  url = {https://qwen.ai/blog?id=qwen3.8},
}

@misc{deepseekv4,
  author = {{DeepSeek AI}},
  title = {{DeepSeek-V4-Pro model card}},
  year = {2026},
  url = {https://huggingface.co/deepseek-ai/DeepSeek-V4-Pro},
}

@misc{longcat20,
  author = {{Meituan}},
  title = {{LongCat-2.0 model card}},
  year = {2026},
  url = {https://huggingface.co/meituan-longcat/LongCat-2.0},
}

@inproceedings{vllm,
  author = {W.~Kwon and others},
  title = {{Efficient Memory Management for Large Language Model Serving with PagedAttention}},
  booktitle = {SOSP},
  year = {2023},
}

@inproceedings{kivi,
  author = {Zirui Liu and Jiayi Yuan and Hongye Jin and Shaochen Zhong and Zhaozhuo Xu and Vladimir Braverman and Beidi Chen and Xia Hu},
  title = {{KIVI: A Tuning-Free Asymmetric 2bit Quantization for KV Cache}},
  booktitle = {ICML},
  year = {2024},
  pages = {32332--32344},
}

@inproceedings{llmint8,
  author = {T.~Dettmers and M.~Lewis and Y.~Belkada and L.~Zettlemoyer},
  title = {{LLM.int8(): 8-bit Matrix Multiplication for Transformers at Scale}},
  booktitle = {NeurIPS},
  year = {2022},
}

@inproceedings{gptq,
  author = {E.~Frantar and S.~Ashkboos and T.~Hoefler and D.~Alistarh},
  title = {{GPTQ: Accurate Post-Training Quantization for Generative Pre-trained Transformers}},
  booktitle = {ICLR},
  year = {2023},
}

@inproceedings{awq,
  author = {J.~Lin and others},
  title = {{AWQ: Activation-aware Weight Quantization for On-Device LLM Compression and Acceleration}},
  booktitle = {MLSys},
  year = {2024},
}

@misc{llamacpp,
  author = {{ggml-org}},
  title = {{llama.cpp: Tensor Encoding Schemes}},
  year = {n.d.},
  url = {https://github.com/ggml-org/llama.cpp/wiki/Tensor-Encoding-Schemes},
}

@article{nvfp4,
  author = {NVIDIA and others},
  title = {{Pretraining Large Language Models with NVFP4}},
  journal = {arXiv preprint arXiv:2509.25149},
  year = {2025},
}

@misc{mxspec,
  author = {{Open Compute Project}},
  title = {{OCP Microscaling Formats (MX) Specification}},
  year = {2023},
  note = {Version 1.0},
  url = {https://www.opencompute.org/documents/ocp-microscaling-formats-mx-v1-0-spec-final-pdf},
}

@misc{nvidiaptx,
  author = {{NVIDIA}},
  title = {{Parallel Thread Execution ISA, block scaling for matrix multiply-accumulate}},
  year = {n.d.},
  url = {https://docs.nvidia.com/cuda/parallel-thread-execution/index.html},
}

@misc{amdcdna4,
  author = {{AMD}},
  title = {{AMD CDNA 4 Architecture White Paper}},
  year = {2025},
  url = {https://www.amd.com/content/dam/amd/en/documents/instinct-tech-docs/white-papers/amd-cdna-4-architecture-whitepaper.pdf},
}

@inproceedings{atom,
  author = {Y.~Zhao and others},
  title = {{Atom: Low-Bit Quantization for Efficient and Accurate LLM Serving}},
  booktitle = {MLSys},
  year = {2024},
}

@inproceedings{qserve,
  author = {Y.~Lin and others},
  title = {{QServe: W4A8KV4 Quantization and System Co-design for Efficient LLM Serving}},
  booktitle = {MLSys},
  year = {2025},
}

@misc{ninfer,
  author = {{NInfer contributors}},
  title = {{NInfer: serving implementation and performance methodology}},
  note = {Revision a754fcf3},
  year = {n.d.},
  url = {https://github.com/Neroued/ninfer/tree/a754fcf3bf524494aa927c2f54e0cbfb58f569ab},
}

@article{oppenheim,
  author = {A.~V.~Oppenheim},
  title = {{Realization of Digital Filters Using Block-Floating-Point Arithmetic}},
  journal = {IEEE Transactions on Audio and Electroacoustics},
  volume = {18},
  number = {2},
  pages = {130--136},
  year = {1970},
}

@inproceedings{vsquant,
  author = {S.~Dai and others},
  title = {{VS-Quant: Per-vector Scaled Quantization for Accurate Low-Precision Neural Network Inference}},
  booktitle = {MLSys},
  year = {2021},
}

@inproceedings{zeroquant,
  author = {Z.~Yao and others},
  title = {{ZeroQuant: Efficient and Affordable Post-Training Quantization for Large-Scale Transformers}},
  booktitle = {NeurIPS},
  year = {2022},
}

@inproceedings{qlora,
  author = {Tim Dettmers and Artidoro Pagnoni and Ari Holtzman and Luke Zettlemoyer},
  title = {{QLoRA: Efficient Finetuning of Quantized LLMs}},
  booktitle = {NeurIPS},
  year = {2023},
}

@inproceedings{sharedmx,
  author = {B.~Rouhani and others},
  title = {{With Shared Microexponents, A Little Shifting Goes a Long Way}},
  booktitle = {ISCA},
  year = {2023},
}

@article{mxpaper,
  author = {B.~Darvish Rouhani and others},
  title = {{Microscaling Data Formats for Deep Learning}},
  journal = {arXiv preprint arXiv:2310.10537},
  year = {2023},
}

@inproceedings{smoothquant,
  author = {G.~Xiao and J.~Lin and M.~Seznec and H.~Wu and J.~Demouth and S.~Han},
  title = {{SmoothQuant: Accurate and Efficient Post-Training Quantization for Large Language Models}},
  booktitle = {ICML},
  year = {2023},
}

@inproceedings{quarot,
  author = {S.~Ashkboos and others},
  title = {{QuaRot: Outlier-Free 4-Bit Inference in Rotated LLMs}},
  booktitle = {NeurIPS},
  year = {2024},
}

@inproceedings{wikitext,
  author = {S.~Merity and C.~Xiong and J.~Bradbury and R.~Socher},
  title = {{Pointer Sentinel Mixture Models}},
  booktitle = {ICLR},
  year = {2017},
}

@misc{llamacppperplexity,
  author = {{ggml-org}},
  title = {{llama.cpp: Perplexity and KL divergence evaluation}},
  year = {n.d.},
  url = {https://github.com/ggml-org/llama.cpp/blob/master/tools/perplexity/README.md},
}

@misc{unslothdynamic,
  author = {{Unsloth AI}},
  title = {{Unsloth Dynamic v2.0 GGUFs}},
  year = {n.d.},
  url = {https://www.unsloth.ai/blog/dynamic-v2},
}

@misc{vllmonline,
  author = {{vLLM contributors}},
  title = {{Online Quantization}},
  year = {n.d.},
  url = {https://docs.vllm.ai/en/stable/features/quantization/online/},
}

@misc{ninferfork,
  author = {{Alphastorm contributors}},
  title = {{NInfer fork with RTX 4090 and 3090 support}},
  year = {n.d.},
  url = {https://github.com/alphastorm/ninfer},
}

@inproceedings{specdecode,
  author = {Y.~Leviathan and M.~Kalman and Y.~Matias},
  title = {{Fast Inference from Transformers via Speculative Decoding}},
  booktitle = {ICML},
  year = {2023},
  pages = {19274--19286},
}

@inproceedings{mtp,
  author = {Fabian Gloeckle and Badr Youbi Idrissi and Baptiste Rozi{\`e}re and David Lopez-Paz and Gabriel Synnaeve},
  title = {{Better \& Faster Large Language Models via Multi-token Prediction}},
  booktitle = {ICML},
  year = {2024},
  pages = {15706--15734},
}

@misc{amperegpu,
  author = {{NVIDIA}},
  title = {{NVIDIA Ampere GA102 GPU Architecture}},
  year = {n.d.},
  note = {Version 2.1},
  url = {https://www.nvidia.com/content/PDF/nvidia-ampere-ga-102-gpu-architecture-whitepaper-v2.1.pdf},
}

@misc{blackwellgpu,
  author = {{NVIDIA}},
  title = {{NVIDIA RTX Blackwell GPU Architecture}},
  year = {n.d.},
  url = {https://images.nvidia.com/aem-dam/Solutions/geforce/blackwell/nvidia-rtx-blackwell-gpu-architecture.pdf},
}

@article{macaq,
  author = {Muller, Lorenz K. and Bich, Philippe and Boretti, Chiara and Chang, Hyun-Min and Zhuang, Jiawei and Cavigelli, Lukas},
  title = {{MACaQ}: Fast, Model Agnostic Calibrated Quantization for Low-Precision Weights and Activations},
  journal = {arXiv preprint arXiv:2611.10649},
  year = {2026}
}
